| | |
|---|---|
| **Suggested title:** | **A Retrospective Analysis using Deep-Learning Models for Prediction of Survival Outcome and Benefit of Adjuvant Chemotherapy in Stage II/III Colorectal Cancer** |
| **Authors** | Xingyu Li[1], Jitendra Jonnagaddala[2], Shuhua Yang[1], Hong Zhang[1*], Xu Steven Xu[3*] |
| | [1] Department of Statistics and Finance, School of Management, University of Science and Technology of China, Hefei, Anhui 230026, China; |
| | [2] School of Population Health, UNSW Sydney, NSW, Australia |
| | [3] Data Science/Translational Research, Genmab Inc., Princeton, New Jersey, USA; |
| **Corresponding author** | |
| | **Hong Zhang**, Department of Statistics and Finance, School of Management, University of Science and Technology of China, Hefei, Anhui 230026, China |
| | E-mail : zhangh@ustc.edu.cn |
| | **Steven Xu,** Data Science/Translational Research, Genmab Inc., Princeton, New Jersey, USA |
| | E-mail: sxu@genmab.com |
| **Running title** | Prediction of Chemotherapy Benefit for Stage II/III CRC |
| **Keywords** | Deep Learning, whole-slide Images, MCO dataset, H&E Image, Adjuvant Chemotherapy, Overall Survival, Colorectal Cancer |
| **Conflict of interest** | The authors declare no potential conflicts of interest. |

**Main text word count:** 4066

**Title length:** 14

**Abstract word count:** 248

**Funding:**

National Natural Science Foundation of China (No. 11771096, 72091212), Anhui Center for Applied Mathematics, and Special Project of Strategic Leading Science and Technology of CAS (No. XDC08010100). Australian National Health and Medical Research Council (No. GNT1192469). Google Cloud Research Credits program with the award GCP19980904.

**Data availability**

The TCGA dataset is publicly available at the TCGA portal (https://portal.gdc.cancer.gov). The public TCGA clinical data is available at the website(https://xenabrowser.net/datapages/). Xception model weights are available at (https://github.com/fchollet/deep-learning-models/releases/download/v0.4/xception_weights_tf_dim_ordering_tf_kernels_notop.h5).

The MCO dataset (whole-slide images) is available from Molecular and Cellular Oncology but restrictions apply to the availability of data, which were used with permission for the current study, and so are not publicly available.

**Code availability**

Source code is available at https://github.com/1996lixingyu1996/CRCNet .

*Key Points*

**Questions:**

Does deep learning models have the ability facilitate the decision for adjuvant chemotherapy?

**Findings:**

We developed a deep-learning based biomarker that can predict patient prognosis and the treatment effect of adjuvant chemotherapy in stage II/III CRC using haematoxylin and eosin images. The biomarker-identified high-risk subgroup significantly benefits from adjuvant chemotherapy whereas no chemotherapy benefit is observed in the low- and medium-risk groups.

**Meaning:**

The biomarker can automate patient selection to deliver "right" adjuvant treatments to "right" Stage II/III CRC patients, improving patient survival, and avoiding unnecessary treatment and associated toxicity.

# Abstract

**IMPORTANCE:** Most of Stage II/III colorectal cancer (CRC) patients can be cured by surgery alone, and only certain CRC patients benefit from adjuvant chemotherapy. Risk stratification based on deep-learning from haematoxylin and eosin (H&E) images has been postulated as a potential predictive biomarker for benefit from adjuvant chemotherapy. However, so far, no direct evidence has been reported.

**OBJECTIVE:** To develop a deep learning-based algorithm (CRCNet) using H&E images to (1) predict the survival outcome in stage II/III CRC, (2) derive risk stratification, and (3) study the association between benefit of adjuvant chemotherapy and risk classes. .

**DESIGN, SETTING, AND PARTICIPANTS:** This deep learning model was conducted from April 1,2021, to July 1, 2021, as a retrospective analysis of data. We trained and internally validated CRCNet using 780 Stage II/III CRC patients from Molecular and Cellular Oncology. Independent external validation of the model was performed using 337 Stage II/III CRC patients from The Cancer Genome Atlas.

**MAIN OUTCOMES AND MEASURES:** The main outcomes are deep learning models using H&E-stained images for risk stratification, prediction of survival and benefit of adjuvant chemotherapy in Stage II/III CRC patients.

**RESULTS:** CRCNet stratified the patients into high, medium, and low-risk subgroups, which accurately and consistently predicts the overall survival in Stage II/III CRC. Multivariate Cox regression analyses confirmed that CRCNet risk groups are statistically significant after adjusting for existing risk factors. The high-risk subgroup significantly benefits from adjuvant chemotherapy. A hazard ratio (chemo-treated vs untreated) of 0.2 (95% CI, 0.05 to 0.65; $P$ = 0.009) and 0.6 (95% CI, 0.42 to 0.98; $P$ = 0.038) are observed in the TCGA and MCO 5FU treated patients, respectively. Conversely, no significant benefit from chemotherapy is observed in the low- and medium-risk groups ($P$ = 0.2 – 1).

**CONCLUSIONS AND RELEVANCE:** The retroepctive analysis provides further evidence that H&E image-based biomarkers may potentially be of great use in delivering treatments following surgery for Stage II/III CRC, improving patient survival, and avoiding unnecessary treatment and associated toxicity, and warrants further validation on other datasets and prospective confirmation in clinical trials.

**Introduction**

Colorectal cancer (CRC) is the second most common cause of cancer-related death, with 1,880,725 new cases and 915,880 deaths worldwide in 2020.[1] For early stage CRC patients, the use of postoperative adjuvant chemotherapy is being debated since surgery is extremely effective, and the majority (approximately 80%) of patients can be cured by surgery alone.[2-4] In addition, significant side effects and an increased death rate have been associated with adjuvant chemotherapy.[5-8] In practice, adjuvant chemotherapy is only recommended to patients with pathological T4 stage tumors, poorly differentiated tumor, vascular, lymphatic, or perineural invasion, lymph nodes sampling less than 12, or clinical presentation with intestinal occlusion or perforation high risk.[9,10]

Recent advancements of artificial intelligence based on deep learning algorithms have greatly improved the current pathological workflows.[11] Skrede et al. developed a prognostic marker using deep learning of H&E slides.[12] This marker could identify high-risk stage II and III CRC patients based on survival. In addition, Kather et al. proposed CNN-based "deep stroma score" (DSS) prognostic factor for overall survival (OS) in Stage III/IV CRC patients.[13] DSS could identify subgroups with more aggressive disease in Stage III/IV CRC patients although its prognostic value was limited in Stage I/II CRC. Danielsen et al. proposed an image-based "tumor-stroma fraction" that could predict prognosis in Stage II CRC.[14] However, a more recent analysis showed that this biomarker was not a significant prognostic factor.[15] Furthermore, Yao et al. proposed an attention-based multiple instance learning model to predict the survival of CRC patients.[16] In addition, Wulczyn et. al. developed a deep learning system (DLS) for predicting disease-specific survival for stage II and III colorectal cancer. Furthermore, Skrede et al. developed a prognostic marker to identify high-risk stage II and III CRC patients using deep learning of H&E slides, and postulated that the identified high-risk CRC could benefit from adjuvant chemotherapy.[12,17] Nevertheless, so far, no analysis has been performed to demonstrate that the model-based risk stratification could predict the treatment effect of adjuvant chemotherapy and identify patients that could benefit from the chemotherapy after surgery. In this study, we use a novel deep-learning model (CRCNet) to integrate predictive imaging phenotypes from different tissue types of whole-slide H&E images (WSIs) from two large international CRC datasets from Molecular and Cellular Oncology (MCO) and The Cancer Genome Atlas (TCGA) studies, and predict prognosis of Stage II/III CRC patients. Our retrospective analysis provided further evidence that the risk stratification based on H&E images could predict the patients that potentially benefit from postoperative adjuvant chemotherapy and provide a useful tool for individualized guidance for treatment and patient care for Stage II/III CRC patients.[18-20]

**Methods**

*Imaging and Clinical Data*

The analysis included two large-scale datasets. Hematoxylin and eosin (H&E) stained whole-slide images (WSIs, 40x) were collected from both MCO and TCGA CRC studies. The MCO dataset consisted of patients who underwent curative resection for colorectal cancer between 1994 to 2010 in New South Wales, Australia. The Cancer Genome Atlas (TCGA) public dataset included the TCGA-COAD and TCGA-READ datasets. The MCO dataset (v15Jan2021) was used to train the deep learning model while the TCGA dataset obtained in July 2020 was used for external validation.

*Deep learning-based risk stratification*

The deep learning model (CRCNet) consisted of two sequential components: a tissue-type classifier and a deep multi-instance learning (MIL) survival model. Each whole-slide H&E image was preprocessed to (1) exclude the background area of each image using a Unet, (2) split into non-overlapping tiles with a size of 224 x 224 pixels, and color normalized. An Xception model-based tissue-type classifier was fine-tuned and classified each image tile into one of eight tissue classes: adipose tissue (ADI), background (BACK), debris (DEB), lymphocytes (LYM), mucus (MUC), smooth muscle (MUS), normal colon mucosa (NORM), cancer-associated stroma (STR), and colorectal adenocarcinoma epithelium (TUM). For each tissue type on the tissue map, a deep MIL survival model was developed based on a feature matrix (tiles x 256) extracted from the last layer of the Xception model. A convolutional one-dimensional layer was used to estimate a score for each tile. The 10 highest and 10 lowest tiles scores of each tissue type were used to predict the patient's risk score. For the MIL model of each tissue type, the patient was classified into high risk or low risk using the median risk score as a threshold. The top 2 models (tissue types) with the highest C-index (tumor and stroma: C-index = 0.61), were integrated to form an ensemble model and to refine the risk stratification into 3 categories: high risk (both tumor and stroma models = high risk); medium risk (either tumor or stroma model = high risk); and low risk (both tumor and stroma models = low risk).

*Statistical analysis*

Univariate and multivariate survival analyses for baseline clinical variables, molecular features, and CRCNet score were performed using Cox proportional hazards models implemented in the R package survival.[21] Log-rank tests were used to evaluate the statistical significance of the difference in survival distributions between subgroups. C-index was used to assess the model predictive performance and compare different models. A Spearman correlation test was performed to assess the significance of correlation between CRCNet risk classification and existing risk factors. The results were internally validated using the MCO dataset with 5-fold cross-validation strategy where folds were stratified based on disease stage. The MCO dataset was split into training (80%) and validation (20%) datasets. External validation was performed using the TCGA database that was kept entirely separate from the training process.

**Results**

*Patient Characteristics in MCO and TCGA*

A total of 1,117 UICC TNM Stage II/III CRC patients are included in the analysis (Table 1): 780 patients are from the MCO dataset, and 337 patients are from the TCGA dataset.[22] The baseline patient characteristics and demographics are similar between the MCO and TCGA studies. The median age of the MCO dataset is 70 years (range: 24 – 99 years) while the median of the TCGA dataset is 67 years (range: 31 – 90 years). 53% and 51% of the MCO and the TCGA patients, respectively, are male. The MCO population consisted of 53% Stage II patients and 47% Stage III patients, whereas there are 54% and 47% Stage II and III patients, respectively, in the TCGA population. In addition, 16% and 21% of subjects from MCO and TCGA, respectively, have an MSI-H status. After a median follow up of 59 months, 245 death events are recorded in the MCO dataset, while the number of deaths is 42 in the TCGA dataset following a median follow-up of 25.1 months. The median OS is 100 months (95% CI 83.3 to infinity) for the TCGA patients while the median survival is not reached for the Stage II/III MCO patients.

*CRCNet Risk Classification*

In the MCO dataset, 288 (37%), 185 (24%), and 307 (39%) of 780 patients are stratified into the low-, medium-, and high-risk groups, respectively, according to the risk scores predicted from tumor and stroma tissue compartments (see Methods). In the TCGA dataset, 133 (39%), 141 (42%), and 63 (19%) of 337 Stage II/III patients are stratified into the low-, medium-, and high-risk groups, respectively.

*Prediction of Overall Survival*

The CRCNet consistently predicts the OS in Stage II/III CRC in both MCO and TCGA datasets (Figure 2). Compared to patients in the low-risk group, patients with both medium and high CRCNet score have significant shorter OS in the MCO dataset (medium risk: HR = 1.67; 95% CI 1.16 to 2.40; $P$ = 0.0056; high risk: HR = 2.41; 95% CI 1.77 to 3.28; $P$ < 0.0001) (Table 2 and Figure 2a). Similarly, the statistical significance is demonstrated in the high-risk group of the TCGA dataset (HR = 2.84; 95% CI 1.26 to 6.40; $P$ = 0.01) (Table 2 and Figure 2b). The univariate analysis indicates the consistency and superiority in prediction performance of CRCNet compared to other reported risk factors such as age, sex, lymphovascular invasion, pT stage, pN stage, UICC stage, KRAS mutation, BRAF mutation as well as MSI status (Table 2).

When stratified by the UICC stage, clear separation of survival between different CRCNet risk groups are still observed in Stage II or III colorectal cancer patients from both MCO and TCGA studies (Figure 2c-f). The CRCNet classifier also identifies significantly different survival within the pT stage (pT1-pT3 or pT4 disease; Supplementary Figure 2). Similarly, CRCNet provides a consistent prediction in the subgroups of pN0 – N1 and pN2 (Supplementary Figure 3), where a substantial difference in OS is identified among different risk groups. In addition, our CRCNet can further separate the risk for MSS patients, and the high-risk MSS patients demonstrate a significantly poorer prognosis (Supplementary Figure 4c and 4d). Within the MSI-H status, the survival is similar irrespective of risk groups (Supplementary Figure 4a and 4b). The predictive ability of OS by CRCNet in subgroups of established risk parameters indicates that features beyond the current clinical and biomarker risk factors can be captured using our deep learning algorithm implemented in CRCNet.

In the multivariate Cox PH models combining CRCNet with other established clinical and molecular risk factors (i.e., pT stage, pN stage, MSI status, lymphovascular invasions, BRAF mutation, KRAS mutation, age, and sex), the CRCNet risk classifier remains statistically significant in both MCO (high risk vs low risk: HR = 2.14; 95% CI, 1.56 to 2.93; $P$ < 0.001) and TCGA datasets (HR = 2.76; 95% CI, 1.17 to 6.50; *P* = 0.02) (Table 3). The statistical significance of CRCNet in the multivariate setting confirms that CRCNet is a robust predictor of OS and provides extra predictive information in addition to the other existing risk parameters in the Stage II/III CRC patients.

More detailed information regarding prediction of overall survival is in Supplementary materials.

*Prediction of Survival Benefit from Adjuvant Chemotherapy*

The classification of high-risk Stage II/III CRC patients using deep learning models may facilitate identification of patients who can benefit from adjuvant chemotherapy and help the treatment decision after surgery. We retrospectively examine the effect of postoperative adjuvant chemotherapy between patients who did or did not receive adjuvant chemotherapy in each CRCNet risk group (Figure 3).

Overall, in the MCO dataset, among the Stage II/III CRC patients, 334 patients received adjuvant chemotherapy while 280 patients did not have chemotherapy. In the TCGA dataset, 111 Stage II/III patients received adjuvant chemotherapy and 135 patients did not have postoperative chemotherapy. However, analysis shows that patients who received postoperative adjuvant chemotherapy do not have a significant survival benefit from the treatment compared to those who did not receive adjuvant chemotherapy in both MCO (HR = 0.9; 95% CI, 0.67 to 1.17; *P* = 0.4; Figure 3) and TCGA (HR = 0.5; 95% CI, 0.26 to 1.12; *P* = 0.1; Figure 3) studies. The 4-year survival is 74.8% with chemotherapy vs 70.8% without chemotherapy in the MCO dataset, while the 4-year survival is 85.8% with chemotherapy vs 77.7% without chemotherapy in the TCGA dataset. This suggests that there is still room for improvement in terms of current clinical decision for adjuvant chemotherapy.

CRCNet-based risk classification clearly predicts the benefit of chemotherapy in the TCGA dataset (Figure 3). The CRCNet low-risk patients who received postoperative chemotherapy do not show any survival benefit compared to the low-risk patients not receiving chemotherapy (HR = 1.1; 95% CI, 0.28 to 4.52; $P$ = 0.87). The CRCNet medium-risk patients who had chemotherapy appear to have a numerical reduction in risk (HR = 0.9; 95% CI, 0.32 to 2.6; $P$ = 0.86), whereas in the CRCNet high-risk subgroup, adjuvant chemotherapy significantly reduces the risk of death (HR = 0.2; 95% CI, 0.05 to 0.65; $P$ = 0.009). No detailed information regarding drug regimens/classes for the adjuvant chemotherapy is provided in the TCGA dataset.

Similar findings are observed in the MCO dataset. The Forest plot analysis (Figure 3) reveals the predictive value of CRCNet for treatment effect of chemotherapy in terms of OS, where the CRCNet risk groups (low- to high-risk) demonstrate a distinct response to chemotherapy. In the low-risk groups, adjuvant chemotherapy does not provide significant benefit of survival (HR = 1.0; 95% CI, 0.57 to 1.77; $P$ = 0.99). Numerical improvement of survival is observed in the chemo-treated medium-risk patients (HR = 0.8; 95% CI, 0.47 to 1.47; $P$ = 0.52). In contrast, in the CRCNet high-risk group, there is a borderline association of chemotherapy with survival (HR = 0.7; 95% CI, 0.46 to 1.01; $P$ = 0.057). Further analysis of the MCO dataset demonstrate that patients in the CRCNet high-risk subgroup benefited from 5FU the most and 5FU is significantly associated with survival benefit (HR = 0.6; 95% CI, 0.42 to 0.98; $P$ = 0.038). In patients who have medium risk according to CRCNet, there is a numerical improvement in survival when receiving 5FU (HR = 0.7; 95% CI, 0.33 to 1.32; $P$ = 0.24).

The chemotherapy-treated patients in both MCO and TCGA data are significantly younger and more fit (73% and 78% < 70 yr in MCO and TCGA, respectively; Supplementary Table 1). In addition, among patients who received adjuvant chemotherapy in both studies, only 8-9% patients are with an MSI-H status, and approximately 90% are MSS patients. Significantly, more pN2 patients are treated with adjuvant chemotherapy in both MCO (27% in treated group vs 7% in untreated) and TCGA (25% in treated vs 9% in untreated). However, even though more pT4 patients are allocated to chemotherapy (28% vs 17% in the untreated group), a similar proportion of pT4 patients are treated (11%) and untreated (7%) groups.

Despite the potential confounding between treatment allocations and clinical risk factors, further Forest plot analysis within the subgroups of these clinical risk factors reveals that the association between CRCNet and adjuvant chemotherapy is not substantially impacted by these known risk factors (Supplementary Figure 6). That is, we perform additional examination of the survival benefit of postoperative chemotherapy for different CRCNet risk groups after stratifying the patients by pT stage, pN stage, MSI status, and lymphovascular invasion. Due to the small sample size in subpopulations, we pool the data from MCO and TCGA together for this analysis. In general, more profound treatment benefit is observed in the CRCNet high-risk group in all subgroups (i.e., pT1-T3/pT4, pN0-pN1/pN2, MSI-H/MSS, age ($\leq$ 70 years / > 70 years), and lymphovascular invasion (yes/no)) compared to those in the CRCNet low- and medium-risk groups. However, it is worth mentioning that MSI-H patients have minimal benefit from chemotherapy even though the CRCNet high-risk, MSI-H patients appears to have a numerically greater benefit from chemotherapy compared to the CRCNet low- and medium-risk, MSI-H patients. This supports the NCCN recommendation that adjuvant chemotherapy is not needed for MSI-H patients.

**Discussion**

The most challenging question in the adjuvant setting for CRC is which patients should receive chemotherapy, as most of the patients can be cured by surgery alone. Currently, the treatment decision is mainly based on clinical and pathological staging. Although novel biomarkers based on genetic mutation status, gene expression profiling, and immunohistochemistry have been developed to facilitate the decision for adjuvant chemotherapy, limited success has been achieved due to moderate prognostic accuracy and/or lack of prediction ability for treatment benefit.[23,24] Skrede et al. indicated that H&E image-based high-risk CRC might benefit better from adjuvant chemotherapy.[12] In this manuscript, we developed a novel deep learning model based on whole-slide H&E images to predict clinical outcome for Stage II/III CRC and provided further evidence that image-based risk stratification may predict the treatment effect of adjuvant chemotherapy.

In both MCO and TCGA studies, the CRCNet identified high-risk subgroups benefited from adjuvant chemotherapy the most, and the OS in these high-risk patients who received chemotherapy (particularly 5FU in the MCO dataset) after surgery was significantly better than that in those who were not given postoperative chemotherapy. In contrast, the CRCNet identified low- and medium-risk patients did not respond to adjuvant chemotherapy or only had numerical improvement with chemotherapy vs. without chemotherapy. In the CRCNet identified high-risk subgroup, 61% patients from the TCGA dataset and 51% from the MCO dataset were not offered any chemotherapy after surgery. Those subjects could have longer survival if they were given adjuvant chemotherapy. On the other hand, those patients who were in the CRCNet low- and medium-risk subgroups and received adjuvant chemotherapy might be able to avoid unnecessary treatment and chemo-associated toxicity. Therefore, our model strongly suggests that H&E image-based biomarkers like CRCNet may be a predictive biomarker that may facilitate selection of high-risk Stage II/III CRC patients who can benefit from adjuvant chemotherapy. However, Due to the retrospective nature of this analysis, further research and validation of H&E image-based biomarkers on other datasets and prospective, randomzied clinical trials is warranted.

Multiple deep-learning models have been developed to use H&E images to predict prognosis of CRC patients.[12,13,16,17] The C-index of these models are all in the range of 0.6 (Supplementary Figure 8). It is well known that the performance of deep-learning models are heavily dependent on sample size (e.g., number of images and number of patients in the dataset). Compared to the models developed by Skrede et al. (Cindex = 0.674) and Wulczyn et. al. (Cindex = 0.66), CRCNet and DeepAttnMISL appears to have lower C-index values (0.62 and 0.606). [12,17] However, after examination of the sample size for each study, the higher C-index values of Skrede et al. and Wulczyn et. al. models were probably due to the larger sample sizes in these two studies instead of difference in the model architectures or workflows. [12,17]

Several assays based on gene expression profiling (Oncotype DX, ColoPrint, and ColDx), immunohistochemistry (e.g., validated Immunoscore according to CD3+ and CD8+ immune cell densities), and post-surgical circulating tumor DNA (ctDNA) have been developed to provide prognostic and predictive information to facilitate the decisions regarding adjuvant therapy in Stage II/III CRC patients.[25-30] However, all these assays may be limited by the complex sample preparation processes, high cost, and issues like large batch effects, etc. In clinical practice, histological slides of tumor tissue, particularly H&E slides, are available for almost every subject. Therefore, image-based approaches can be a great alternative to genomic, transcriptomic, or immunohistochemistry assays due to its accessibility and low cost. Furthermore, automation of image-based workflows largely eliminates human involvement, shortens the image processing time, and allows quick turnaround for clinical decisions.

Taken together, we developed a CNN-based deep learning model on whole-slide images from CRC patients. This model integrates a supervised tissue type classification and a deep-learning approach to extract prognostic features from small H&E image tiles, which allows for prediction of survival of Stage II/III CRC patients and identification of high-risk Stage II/III CRC patients who can potentially benefit from adjuvant chemotherapy. The retrospective analysis provides further evidence that coupling deep-learning models and H&E images can potentially reveal novel predictive biomarkers that offer better decision making for therapeutic allocations for Stage II/III CRC patients.

## Acknowledgements

The research of Xingyu Li, Shuhua Yang, and Hong Zhang was partially supported by National Natural Science Foundation of China (No. 11771096, 72091212), Anhui Center for Applied Mathematics, and Special Project of Strategic Leading Science and Technology of CAS (No. XDC08010100). Jitendra Jonnagaddala is funded by the Australian National Health and Medical Research Council (No. GNT1192469). This dataset was conducted as part of the Translational Cancer research network (TCRN) research program. TCRN was funded by Cancer Institute of New South Wales and Prince of Wales Clinical School, UNSW Medicine. This material is based upon work supported by the Google Cloud Research Credits program with the award GCP19980904. We thank Michelle Xu (Princeton Day School) for writing assistance and language editing.

## Author contributions

X.S.X., X.L., and H.Z. contributed to design of the research; J.J., X.L., and X.S.X. contributed to data acquisition; X.L., X.S.X., and S.Y. contributed to data analysis. X.L., X.S.X., J.J., and H.Z. contributed to data interpretation. X.L., X.S.X., J.J., and H.Z. wrote the manuscript; and all authors critically reviewed the manuscript and approved the final version.

## Competing interests

The authors declare no potential conflicts of interest

**Data availability**

The TCGA dataset is publicly available at the TCGA portal (https://portal.gdc.cancer.gov). The public TCGA clinical data is available at the website(https://xenabrowser.net/datapages/). Xception model weights are available at (https://github.com/fchollet/deep-learning-models/releases/download/v0.4/xception_weights_tf_dim_ordering_tf_kernels_notop.h5).

The MCO dataset (whole-slide images) is available from Molecular and Cellular Oncology but restrictions apply to the availability of data, which were used with permission for the current study, and so are not publicly available.

**Code availability**

Source code is available at https://github.com/1996lixingyu1996/CRCNet .

Figure and Table Legends

**Figure 1. Flow-chart and global methodology of the study.**

The deep learning model (CRCNet) consisted of two sequential components: a tissue-type classifier and a deep multi-instance learning (MIL) survival model. Each whole-slide H&E image was preprocessed to (1) exclude the background area of each image using a Unet, (2) split into non-overlapping tiles with a size of 224 x 224 pixels, and color normalized. An Xception model-based tissue-type classifier was fine-tuned and classified each image tile into one of eight tissue classes: adipose tissue (ADI), background (BACK), debris (DEB), lymphocytes (LYM), mucus (MUC), smooth muscle (MUS), normal colon mucosa (NORM), cancer-associated stroma (STR), and colorectal adenocarcinoma epithelium (TUM). For each tissue type on the tissue map, a deep MIL survival model was developed based on a feature matrix (tiles x 256) extracted from the last layer of the Xception model. A convolutional one-dimensional layer was used to estimate a score for each tile. The 10 highest and 10 lowest tiles scores of each tissue type were used to predict the patient's risk score. For the MIL model of each tissue type, the patient was classified into high risk or low risk using the median risk score as a threshold. The top 2 models (tissue types) with the highest C-index (tumor and stroma: C-index = 0.61), were integrated to form an ensemble model and to refine the risk stratification into 3 categories: high risk (both tumor and stroma models = high risk); medium risk (either tumor or stroma model = high risk); and low risk (both tumor and stroma models = low risk).

**Figure 2. Kaplan Meier plot for risk subgroups according CRCNet for survival in colorectal cancer patients (a) MCO Stage II/III, (b) TCGA Stage II/III, (c) MCO Stage II, (d) TCGA Stage II, (e) MCO Stage III, and (f) TCGA Stage III.** Molecular and Cellular Oncology (MCO); The Cancer Genome Atlas (TCGA).

**Figure 3. Forest plot showing the effect of adjuvant chemotherapy on overall survival among patients with different risk categories according to CRCNet.**

**Table 1. Baseline patient characteristics of the training (MCO) and validation (TCGA) dataset populations**

**Table 2. Univariate analysis of overall survival in MCO and TCGA cohorts**

**Table 3. Multivariate Cox Regression Model in MCO and TCGA Studies**

**Figure 1.**

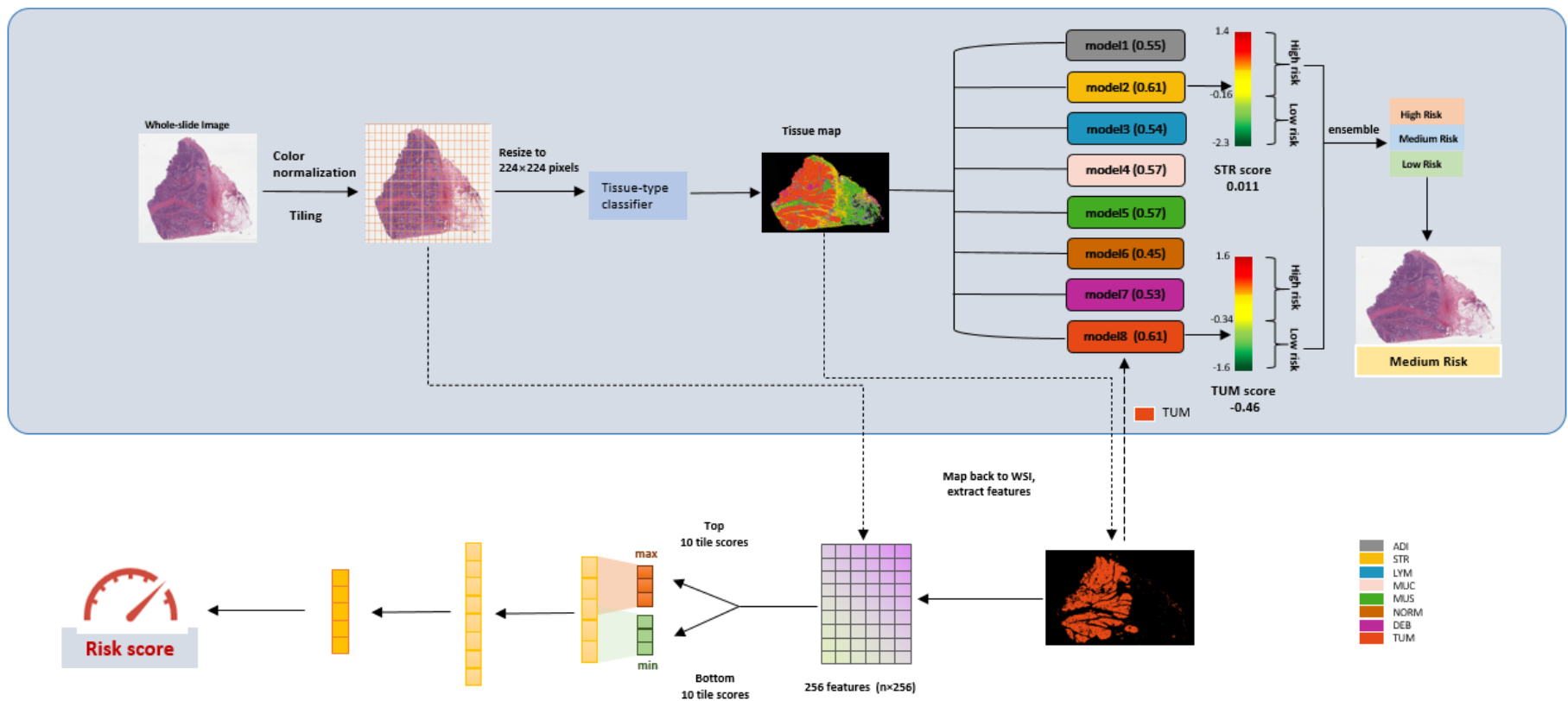

**Figure 2.**

a
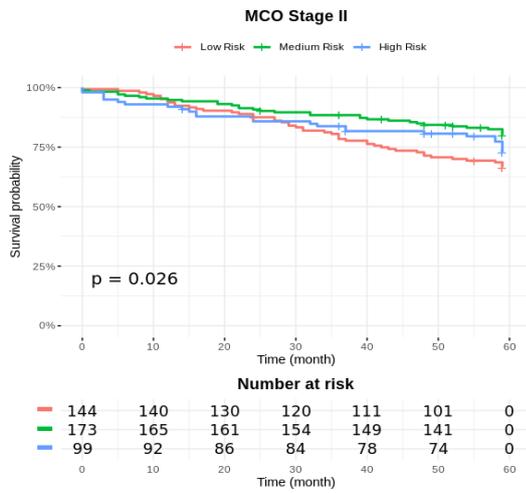

b
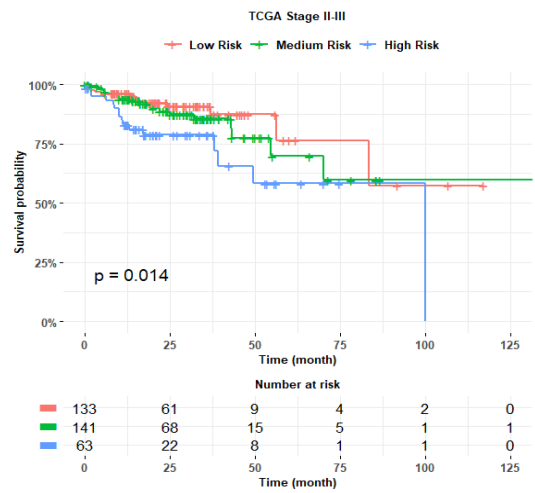

c
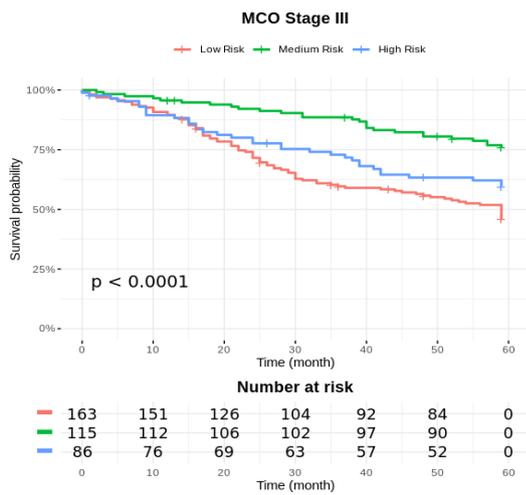

d
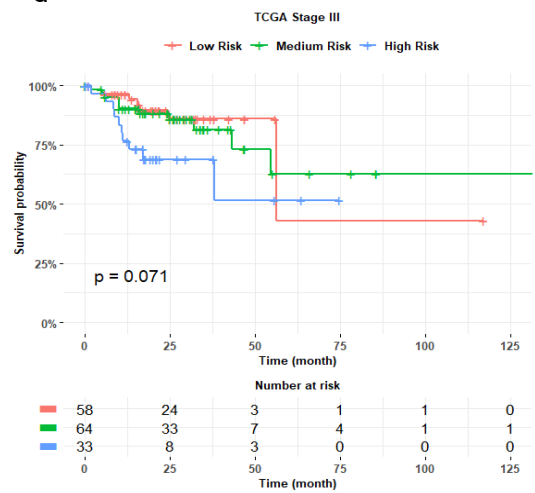

e
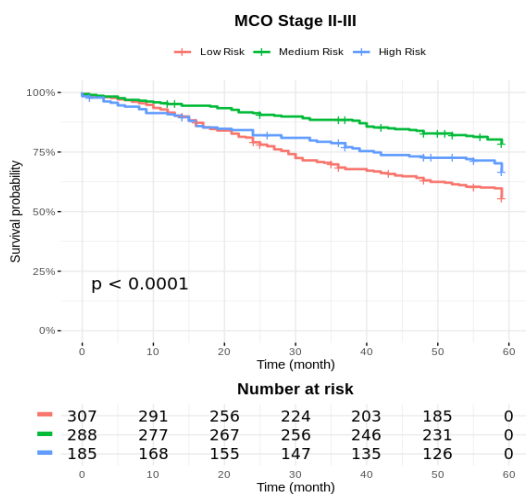

f
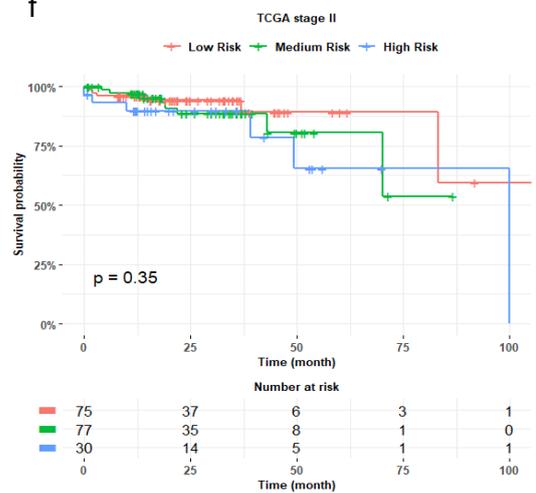

**Figure 3.**

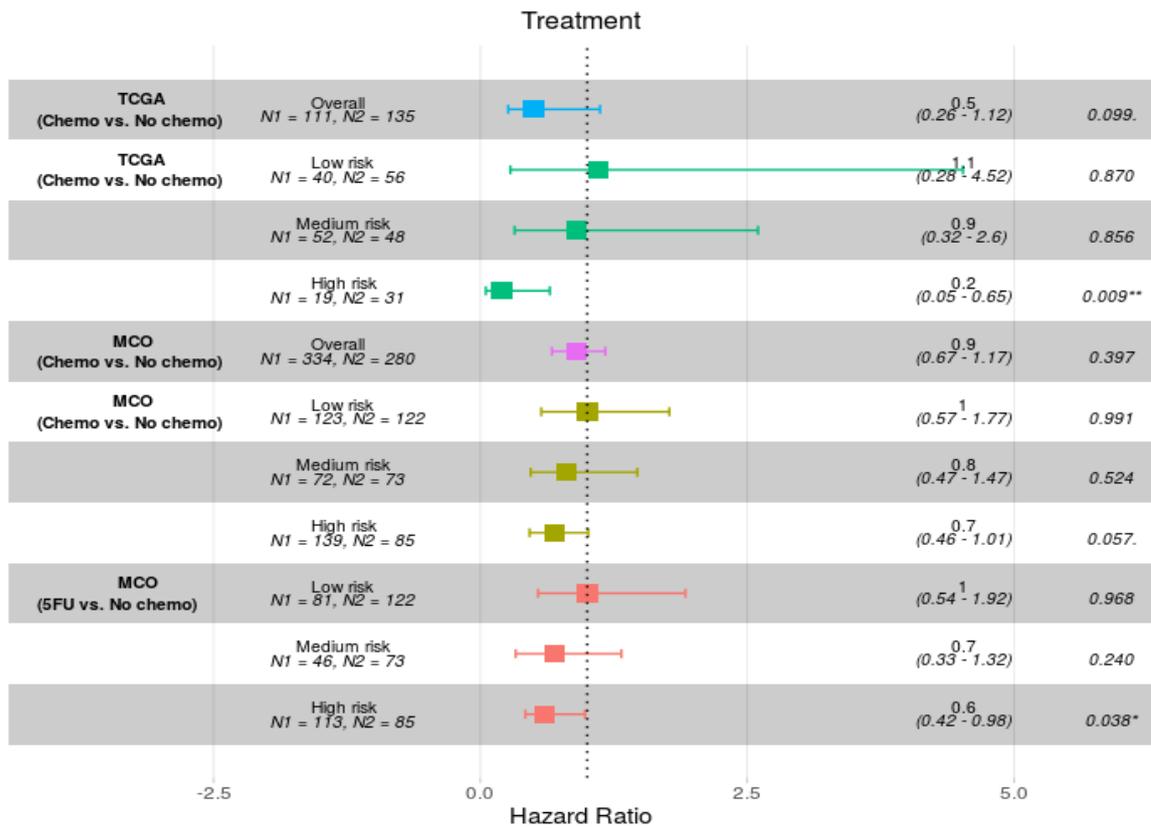

Significant log rank P value:

(.) 0.1 ≥ P > 0.05; (*) 0.05 ≥ P > 0.01; (**) 0.01 ≥ P >0.001; (***) 0.001 ≥ P.

**Table 1.**

|  | MCO cohort | TCGA cohort |
|---|---|---|
|  | (n = 780) | (n = 337) |
| Age, years | 70 (24-99) | 66 (31-90) |
| **Sex** | | |
|    Female | 361(46%) | 166(49%) |
|    Male | 413(53%) | 171(51%) |
|    Missing | 6(1%) | -- |
| **Stage** | | |
|    II | 416(53%) | 182(54%) |
|    III | 364(47%) | 155(46%) |
| **pN stage** | | |
|    N0-N1 | 647(83%) | 283(84%) |
|    N2 | 127(165) | 54(16%) |
|    Missing | 6(1%) | -- |
| **pT stage** | | |
|    pT1-pT3 | 595(76%) | 306(91%) |
|    pT4 | 179(23%) | 31(9%) |
|    Missing | 6(1%) | -- |
| **Location** | | |
|    Colon | 501(64%) | 254(75%) |
|    Rectum | 279(36%) | 83(25%) |
| **Adjuvant treatment** | | |
|    No treatment | 466(60%) | 165(49%) |
|    Treatment | 470(60%) | 136(40%) |
|    5 Fu | 305(39%) | -- |
|    Other treatment | 165(21%) | -- |
|    Missing | 243(31%) | 36(11%) |
| **MSI** | | |
|    MSI-H | 124(16%) | 59(18%) |
|    MSS | 642(82%) | 246(73%) |
|    Missing | 14(2%) | 32(9%) |
| **KRAS** | | |
|    Wild type | 511(66%) | 168(50%) |
|    Mutated | 251(32%) | 106(31%) |
|    Missing | 18(2%) | 63(19%) |
| **BRAF** | | |
|    Wild type | 659(84%) | 237(70%) |
|    Mutated | 101(13%) | 37(11%) |
|    Missing | 20(3%) | 63(19%) |
| **Venous invasion** | | |
|    Absent | 585(75%) | 226(67%) |
|    Present | 195(25%) | 74(22%) |
|    Missing | -- | 37(11%) |
| **Lymphatic invasion** | | |
|    Absent | 474(61%) | 174(52%) |
|    Present | 306(39%) | 145(43%) |
|    Missing | -- | 18(5%) |
| Data are median(range) or n(%) | | |

.

**Table 2**

| Univariate Cox Regression Model in MCO and TCGA Studies | | | | | | |
|---|---|---|---|---|---|---|
| | MCO | | | TCGA | | |
| Variable | HR(95% CI) | P value | C-index(SD) | HR(95% CI) | P value | C-index(SD) |
| CRCNet Score | -- | -- | 0.599(0.017) | -- | -- | 0.621(0.046) |
|   Low | 1(ref) | -- | -- | 1(ref) | -- | -- |
|   Medium | 1.67(1.16,2.40) | 0.006** | -- | 1.70(0.78,3.69) | 0.180 | -- |
|   High | 2.41(1.77,3.28) | <0.0001*** | -- | 2.84(1.26,6.40) | 0.012* | -- |
| KRAS | -- | -- | 0.521(0.015) | -- | -- | 0.539(0.041) |
|   Mutated | 1(ref) | -- | -- | 1(ref) | -- | -- |
|   Wild | 0.84(0.65,1.09) | 0.192 | -- | 1.51(0.77,2.95) | 0.229 | -- |
| BRAF | -- | -- | 0.508(0.011) | -- | -- | 0.500(0.026) |
|   Mutated | 1(ref) | -- | -- | 1(ref) | -- | -- |
|   Wild | 1.17(0.80,1.73) | 0.424 | -- | 1.17(0.80,1.73) | 0.203 | -- |
| MSI Status | -- | -- | 0.514(0.012) | -- | -- | 0.455(0.027) |
|   MSS | 1(ref) | -- | -- | 1(ref) | -- | -- |
|   MSI-H | 1.29(0.89,1.86) | 0.174 | -- | 0.98(0.48,1.99) | 0.947 | -- |
| Sex | -- | -- | 0.539(0.016) | -- | -- | 0.491(0.044) |
|   Male | 1(ref) | -- | -- | 1(ref) | -- | -- |
|   Female | 0.75(0.58,0.96) | 0.024* | -- | 1.06(0.58,1.94) | 0.854 | -- |
| Age | 1.04(1.02,1.05) | <0.0001*** | 0.615(0.018) | 1.02(0.99,1.04) | 0.171 | 0.566(0.051) |
| Lymphovascular Invasion | -- | -- | 0.571(0.015) | -- | -- | 0.599(0.043) |
|   No | 1(ref) | -- | -- | 1(ref) | -- | -- |
|   Yes | 2.00(1.54,2.60) | <0.0001*** | -- | 1.90(1.01,3.58) | 0.047* | -- |
| pT stage | -- | -- | 0.567(0.015) | -- | -- | 0.586(0.038) |
|   T1-T3 | 1(ref) | -- | -- | 1(ref) | -- | -- |
|   T4 | 1.95(1.49,2.55) | <0.0001*** | -- | 2.83(1.30,6.18) | 0.009** | -- |
| pN stage | -- | -- | 0.567(0.013) | -- | -- | 0.616(0.041) |
|   N1 | 1(ref) | -- | -- | 1(ref) | -- | -- |
|   N2 | 2.31(1.73,3.06) | <0.0001*** | -- | 2.79(1.44,5.38) | 0.002** | -- |
| Stage | -- | -- | 0.566(0.016) | -- | -- | 0.591(0.042) |
|   Stage II | 1(ref) | -- | -- | 1(ref) | -- | -- |
|   Stage III | 1.68(1.31,2.17) | <0.0001*** | -- | 1.74(0.94,3.23) | 0.079. | -- |

**Table 3.**

| Multivariate Cox Regression Model in MCO and TCGA Studies | | | | |
|---|---|---|---|---|
| | **MCO** | | **TCGA** | |
| **Variable** | **HR(95% CI)** | **P value** | **HR(95% CI)** | **P value** |
| **CRCNet Score** | | | | |
|   Low | 1(ref) | -- | 1(ref) | -- |
|   Medium | 1.77(1.22,2.55) | 0.002** | 1.89(0.85,4.20) | 0.117 |
|   High | 2.14(1.56,2.93) | <0.0001*** | 2.76(1.17,6.50) | 0.020* |
| **KRAS** | | | | |
|   Mutated | 1(ref) | -- | 1(ref) | -- |
|   Wild | 0.84(0.63,1.10) | 0.201 | 1.66(0.81,3.41) | 0.166 |
| **BRAF** | | | | |
|   Mutated | 1(ref) | -- | 1(ref) | -- |
|   Wild | 0.90(0.55,1.47) | 0.682 | 0.92(0.34,2.44) | 0.860 |
| **MSI Status** | | | | |
|   MSS | 1(ref) | -- | 1(ref) | -- |
|   MSI-H | 1.20(0.76,1.90) | 0.442 | 1.41(0.58,3.43) | 0.454 |
| **Sex** | | | | |
|   Male | 1(ref) | -- | 1(ref) | -- |
|   Female | 0.69(0.53,0.90) | 0.005** | 1.10(0.58,2.10) | 0.767 |
| **Age** | 1.04(1.03,1.05) | <0.0001*** | 1.02(1.00,1.05) | 0.074. |
| **Lymphovascular Invasion** | | | | |
|   No | 1(ref) | -- | 1(ref) | -- |
|   Yes | 1.75(1.33,2.30) | <0.0001*** | 1.36(0.68,2.73) | 0.390 |
| **pT stage** | | | | |
|   T1-T3 | 1(ref) | -- | 1(ref) | -- |
|   T4 | 1.58(1.20,2.08) | 0.001** | 3.10(1.37,7.03) | 0.007** |
| **pN stage** | | | | |
|   N0-N1 | 1(ref) | -- | 1(ref) | -- |
|   N2 | 1.95(1.46,2.62) | <0.0001*** | 2.28(1.09,4.79) | 0.029* |

Note: C-index of the MCO multivariate model = 0.715; C-index of the TCGA multivariate model = 0.709